\documentclass{article}

\usepackage{iclr2026_conference}
\iclrfinalcopy  

\usepackage{times}
\usepackage{hyperref}
\usepackage{url}

\usepackage{graphicx}
\usepackage{subcaption}

\usepackage{amsmath}
\usepackage{amssymb}
\usepackage{amsfonts}
\usepackage{mathtools}
\usepackage{bm}

\usepackage{booktabs}
\usepackage{algorithm}
\usepackage{algorithmic}
\usepackage{multirow}
\usepackage{tikz}
\usetikzlibrary{positioning, arrows.meta, calc}

\usepackage{amsmath,amsfonts,bm}

\def\eqref#1{equation~\ref{#1}}

\def\1{\bm{1}}

\DeclareMathAlphabet{\mathsfit}{\encodingdefault}{\sfdefault}{m}{sl}
\SetMathAlphabet{\mathsfit}{bold}{\encodingdefault}{\sfdefault}{bx}{n}

\hypersetup{
    colorlinks=true,
    linkcolor=blue,
    citecolor=blue,
    urlcolor=blue
}

\title{Tiny Recursive Reasoning with Mamba-2 Attention Hybrid}

\author{Wenlong Wang \& Fergal Reid \\
Intercom \\
Dublin, Ireland \\
\texttt{\{wenlong.wang,fergal.reid\}@intercom.io}
}

\begin{document}

\maketitle

\begin{abstract}
Recent work on recursive reasoning models like TRM demonstrates that tiny networks (7M parameters) can achieve strong performance on abstract reasoning tasks through latent recursion---iterative refinement in hidden representation space without emitting intermediate tokens. This raises a natural question about operator choice: Mamba-2's state space recurrence is itself a form of iterative refinement, making it a natural candidate for recursive reasoning---but does introducing Mamba-2 into the recursive scaffold preserve reasoning capability? We investigate this by replacing the Transformer blocks in TRM with Mamba-2 hybrid operators while maintaining parameter parity (6.83M vs 6.86M parameters). On ARC-AGI-1, we find that the hybrid improves pass@2 (the official metric) by +2.0\% (45.88\% vs 43.88\%) and consistently outperforms at higher K values (+4.75\% at pass@100), whilst maintaining pass@1 parity. This suggests improved candidate coverage---the model generates correct solutions more reliably---with similar top-1 selection. Our results validate that Mamba-2 hybrid operators preserve reasoning capability within the recursive scaffold, establishing SSM-based operators as viable candidates in the recursive operator design space and taking a first step towards understanding the best mixing strategies for recursive reasoning.
\end{abstract}

\section{Introduction}

Large language models have excelled in reasoning tasks such as mathematics, code generation, and logical inference through chain-of-thought prompting \citep{wei2022cot}, iterative refinement, and sampling with aggregation \citep{wang2022selfconsistency, yao2023tot}. A growing body of literature suggests that the \emph{recursive process}---rather than model scale alone---is central to reasoning capability \citep{saunshi2025looped, feng2023cottheory}. This represents a fundamental shift from ``bigger models'' to ``more thinking time,'' where performance scales with inference-time computation rather than parameter count alone.

Most compute-scaling approaches operate in token space, producing visible intermediate steps that can be inspected and verified \citep{wei2022cot, nye2021scratchpads}. However, chain-of-thought models often expend computational resources on tokens unrelated to reasoning---for instance, tokens serving grammatical or stylistic purposes. Two key questions remain open: whether this is the most efficient approach, and whether non-reasoning tokens serve any purpose beyond providing additional forward passes for the model to ``ponder'' before producing an answer. The purported visibility and interpretability of explicit reasoning traces is also challenged by questions of faithfulness: the stated reasoning may not reflect the model's actual computation \citep{lanham2023faithfulness}.

An emerging alternative is \emph{latent recursion}: iterative refinement in hidden representation space without emitting intermediate tokens \citep{geiping2025recurrent, saunshi2025looped, hao2024coconut}. The Tiny Recursive Model (TRM) \citep{yang2025trm} and Hierarchical Recursive Model (HRM) \citep{wang2025hierarchical} exemplify this approach by addressing the challenging ARC-AGI dataset with extremely tiny models. Specifically, TRM achieves 44.6\% on ARC-AGI-1 with only 7M parameters through repeated latent updates. This dramatically outperforms comparably-sized models that lack recursive depth, and notably surpasses many commercial LLM APIs on this benchmark. What enables such small recursive models to succeed where large-scale models struggle remains not fully understood, but evidence increasingly points to the recursive process itself as the key ingredient.

TRM and similar recursive reasoning models assume attention-heavy Transformer blocks as the per-step operator. This raises a key question: can alternative operators---particularly those with inherent recurrence like state space models---enter the \emph{design space of recursive reasoning} without degrading capability? Recent literature demonstrates that Mamba-2 hybrid models serve as strong candidates for language tasks \citep{gu2024mamba, dao2024mamba2, lieber2024jamba}, and are particularly promising for test-time computation due to their efficient inference speed \citep{zhang2025m1}. Mamba-2's state space recurrence ($h_t = a_t h_{t-1} + B_t x_t$) is itself a form of iterative refinement, making it a natural and potentially more efficient substrate for recursive reasoning. Validating that such operators preserve reasoning capability within the recursive scaffold is a necessary first step before exploring deeper questions---such as whether the recursive loop can be internalised into SSM state updates, leveraging Mamba's inherent inner recurrence rather than relying solely on outer-loop iteration.

We present a TRM variant where Transformer blocks are replaced with a Mamba-2 + attention hybrid operator \citep{gu2024mamba, dao2024mamba2}, parameter-matched to the original. Our contributions are:
\begin{itemize}
    \item \textbf{C1}: We are the first Mamba-hybrid model for recursive latent reasoning.
    \item \textbf{C2}: Empirical validation on ARC-AGI-1 showing improved pass@2 performance (+2.0\%), with supplementary results on Sudoku and Maze demonstrating competitive accuracy.
    \item \textbf{C3}: Analysis of pass@K patterns revealing a coverage-vs-selection trade-off: the hybrid improves candidate diversity while maintaining top-1 selection quality.
\end{itemize}

\section{Background}

\subsection{TRM: Recursive Reasoning with Tiny Networks}

The Hierarchical Reasoning Model (HRM) \citep{wang2025hierarchical} first demonstrated that extremely small models could achieve remarkable performance on abstract reasoning through latent recursion. With only 27M parameters, HRM achieved 40.3\% on ARC-AGI-1---dramatically outperforming comparably-sized models and notably surpassing many commercial LLM APIs on this benchmark. This result challenged conventional wisdom that reasoning capability requires massive scale.

Building on HRM's success, the Tiny Recursive Model (TRM) \citep{yang2025trm} proposed a simplified architecture that achieves even stronger performance with fewer parameters (5--7M). TRM maintains two latent states: $z_H$ (high-level) and $z_L$ (low-level), updated through $H$ outer cycles and $L$ inner cycles:
\begin{align}
    z_L^{(t+1)} &= f(z_L^{(t)}, z_H^{(t)} + \text{embed}(x)) \\
    z_H^{(t+1)} &= f(z_H^{(t)}, z_L^{(T_L)})
\end{align}
where $f$ is the learned update function (a stack of Transformer blocks). A key architectural difference: HRM uses two separate models to process $z_L$ and $z_H$, whilst TRM finds that a single shared model suffices, enabling the dramatic parameter reduction whilst improving performance to 44.6\% on ARC-AGI-1.

\subsection{ARC-AGI Evaluation Protocol}

The Abstraction and Reasoning Corpus (ARC) \citep{chollet2019arc} evaluates abstract reasoning through visual puzzles. The evaluation protocol used in TRM expands each test input into $\sim$880 augmentations via dihedral transformations and colour permutations. The model runs once per augmentation, predictions are inverse-transformed back to the original space, and results are aggregated by (vote count, average confidence). The pass@K metric measures whether the correct answer appears in the top-K ranked predictions.

\textbf{Official metric}: ARC-AGI uses pass@2 as the primary measurement system to account for tasks with explicit ambiguity requiring two guesses, and to catch unintentional ambiguity or dataset mistakes \citep{arcprize2025}. More broadly, pass@K measures \emph{candidate-set coverage}---whether the model generates the correct answer among its diverse predictions---whilst lower K values (especially pass@1) reflect \emph{winner selection}---whether the aggregation correctly ranks the answer first.

\subsection{Mamba-2 as an Alternative Mixer}

State Space Models (SSMs) offer an alternative to attention for sequence modeling, addressing the quadratic complexity of Transformers. Mamba \citep{gu2024mamba} introduces \emph{selective} SSMs, where model parameters vary with input, enabling the model to selectively propagate or forget information. The core mechanism processes sequences through a recurrent state update:
\begin{equation}
    h_t = \bar{A}_t h_{t-1} + \bar{B}_t x_t
\end{equation}
where $h_t$ is the hidden state, $x_t$ is the input, and $\bar{A}_t$, $\bar{B}_t$ are input-dependent parameters. Mamba-2 \citep{dao2024mamba2} simplifies this to $h_t = a_t h_{t-1} + B_t x_t$ (where $a_t$ is a scalar), achieving 2--8$\times$ faster training through improved hardware utilization via Structured State Space Duality (SSD).

From an operator viewpoint, TRM repeatedly applies an update function to refine latent states. The original TRM uses bidirectional attention for cross-position communication. Mamba-2 provides an alternative mixer with linear complexity. While Mamba excels at sequential dependencies, it processes information causally in one direction. We therefore explore \emph{hybrid} designs that combine Mamba-2's efficient sequential processing with explicit cross-position mixing (attention or dense MLP) to capture bidirectional dependencies.

\section{Method}

\subsection{Architecture: TRM with Hybrid Update Operator}

We preserve TRM's recursive structure while replacing the per-step operator. The recursion schedule remains unchanged: $H_{\text{cycles}}=3$ outer loops and $L_{\text{cycles}}=4$--$6$ inner loops, with the same state representation ($z_H$, $z_L$) and output heads (LM prediction + Q-halt signal for adaptive computation).

\begin{figure}[t]
\centering

\begin{subfigure}{\textwidth}
\centering
\begin{tikzpicture}[
    node distance=2.5cm,
    mamba_block/.style={rectangle, draw, fill=green!20, text width=2cm, text centered, rounded corners, minimum height=1cm, minimum width=2cm},
    attn_block/.style={rectangle, draw, fill=blue!20, text width=2cm, text centered, rounded corners, minimum height=1cm, minimum width=2cm},
    mlp_block/.style={rectangle, draw, fill=orange!20, text width=2cm, text centered, rounded corners, minimum height=1cm, minimum width=2cm},
    arrow/.style={->, >=stealth, thick}
]

\node (input) {Input};
\node[mamba_block, right of=input] (mamba1) {Mamba-2};
\node[mamba_block, right of=mamba1] (mamba2) {Mamba-2};
\node[attn_block, right of=mamba2] (attn) {Attention};
\node[mlp_block, right of=attn] (mlp) {MLP};
\node[right of=mlp] (output) {Output};

\draw[arrow] (input) -- (mamba1);
\draw[arrow] (mamba1) -- node[above=12pt] {\small norm\&add} (mamba2);
\draw[arrow] (mamba2) -- node[above=12pt] {\small norm\&add} (attn);
\draw[arrow] (attn) -- node[above=12pt] {\small norm\&add} (mlp);
\draw[arrow] (mlp) -- node[above=12pt] {\small norm\&add} (output);

\draw[arrow] ($(mlp)!0.5!(output)$) -- ++(0,-1.0) -| ($(input)!0.3!(mamba1)$);

\end{tikzpicture}
\caption{TR-mamba2attn: Mamba-2 blocks followed by Attention and MLP.}
\label{fig:tr_mamba2attn}
\end{subfigure}

\vspace{1cm}

\begin{subfigure}{\textwidth}
\centering
\begin{tikzpicture}[
    node distance=2.5cm,
    attn_block/.style={rectangle, draw, fill=blue!20, text width=2cm, text centered, rounded corners, minimum height=1cm, minimum width=2cm},
    mlp_block/.style={rectangle, draw, fill=orange!20, text width=2cm, text centered, rounded corners, minimum height=1cm, minimum width=2cm},
    arrow/.style={->, >=stealth, thick}
]

\node (input) {Input};
\node[attn_block, right of=input] (attn1) {Attention};
\node[mlp_block, right of=attn1] (mlp1) {MLP};
\node[attn_block, right of=mlp1] (attn2) {Attention};
\node[mlp_block, right of=attn2] (mlp2) {MLP};
\node[right of=mlp2] (output) {Output};

\draw[arrow] (input) -- (attn1);
\draw[arrow] (attn1) -- node[above=12pt] {\small norm\&add} (mlp1);
\draw[arrow] (mlp1) -- node[above=12pt] {\small norm\&add} (attn2);
\draw[arrow] (attn2) -- node[above=12pt] {\small norm\&add} (mlp2);
\draw[arrow] (mlp2) -- node[above=12pt] {\small norm\&add} (output);

\draw[arrow] ($(mlp2)!0.5!(output)$) -- ++(0,-1.0) -| ($(input)!0.3!(attn1)$);

\end{tikzpicture}
\caption{TRM-Att: the original TRM attention model.}
\label{fig:trm_attn}
\end{subfigure}

\caption{Architecture comparison: (a) TR-mamba2attn with Mamba-2 hybrid operator and (b) TRM-attn with attention-based operator. Both use post-norm residual connections (norm\&add) between components.}
\label{fig:architecture_comparison}
\end{figure}
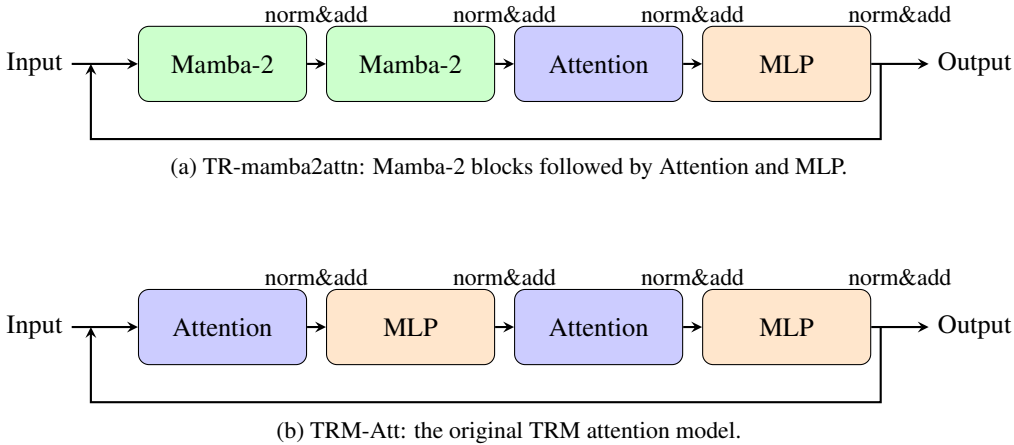

\textbf{What changes}: We swap the Transformer blocks for hybrid block stacks. Figure~\ref{fig:architecture_comparison} illustrates the architectural differences between the original TRM and our hybrid variants. We experiment with two variants:
\begin{itemize}
    \item \textbf{TR-mamba2attn}: As shown in Figure~\ref{fig:architecture_comparison}, this variant replaces the attention-only blocks with a Mamba-2 $\rightarrow$ Mamba-2 $\rightarrow$ Attention $\rightarrow$ MLP pipeline, combining sequential state space processing with cross-position attention.
    \item \textbf{TR-mamba2mlpt}: Similar to TR-mamba2attn, but replaces the attention block with MLP-t, which operates on the transposed sequence dimension for all-to-all cross-position communication without attention.
\end{itemize}

We did not use a pure Mamba model because Mamba's sequential processing is causal by nature, processing information in one direction. For sequence-to-sequence tasks like Sudoku, Maze, and ARC-AGI, bidirectional processing is essential to capture dependencies across the entire spatial grid. For simplicity, we use attention and MLP-t blocks to provide the necessary cross-position information flow, complementing Mamba-2's sequential processing capabilities.

\subsection{Parameter Matching}

To isolate the effect of the operator choice, we match parameters:
\begin{itemize}
    \item Hidden size: 512
    \item TRM-attn: 6.83M parameters
    \item TR-mamba2attn: 6.86M parameters
    \item Mamba-2 specifics: $d_{\text{state}}=128$, $\text{headdim}=64$, $\text{expand}=2$
\end{itemize}

\subsection{Normalisation Choice: Why Post-Norm Matters}

A critical implementation detail for recursive models is the normalisation placement. Modern LLMs typically use \textbf{pre-norm}:
\begin{equation}
    h_{t+1} = h_t + F(\text{Norm}(h_t))
\end{equation}
However, in unrolled recursion where the same residual module is applied $T$ times, pre-norm can allow the residual stream magnitude to grow with $t$ (approximately $\sqrt{t}$ or worse if updates are correlated), eventually causing NaN failures.

\textbf{Post-norm} resolves this by re-normalising after each residual add:
\begin{equation}
    h_{t+1} = \text{Norm}(h_t + F(h_t))
\end{equation}
This bounds the hidden state scale regardless of recursion depth. HRM \citep{wang2025hierarchical} originally motivated post-norm for Q-learning convergence stability in their Adaptive Computation Time mechanism. TRM \citep{yang2025trm} removed the Q-learning objective (finding its contribution to generalisation marginal) whilst retaining post-norm in its architecture, but did not explicitly discuss why. We argue that post-norm is essential for recursion stability itself, independent of Q-learning: it bounds the residual stream magnitude across recursive unrolling, preventing divergence. We follow this design and use post-norm (RMSNorm) in all our experiments.

\section{Experiments}

\subsection{Experimental Setup}

\textbf{Benchmarks}: We evaluate on three tasks from the TRM benchmark suite:
\begin{itemize}
    \item \textbf{ARC-AGI-1}: Abstract reasoning puzzles with up to 30$\times$30 grids, 12-token vocabulary
    \item \textbf{Sudoku-Extreme}: 9$\times$9 constraint satisfaction puzzles
    \item \textbf{Maze-30$\times$30-Hard}: Path-finding through 30$\times$30 mazes
\end{itemize}

\textbf{Models}: We compare four variants:
\begin{itemize}
    \item \textbf{TRM-attn}: Original attention-based TRM (6.83M params)
    \item \textbf{TRM-mlp-t}: TRM with MLP-t blocks (5M params on Sudoku, 19M on ARC/Maze)
    \item \textbf{TR-mamba2attn}: Mamba-2 + Attention hybrid (6.86M params)
    \item \textbf{TR-mamba2mlpt}: Mamba-2 + MLP-t hybrid (6.28M params on Sudoku, 13.24M on ARC/Maze)
\end{itemize}

\textbf{Evaluation}: Sudoku uses exact accuracy; ARC uses pass@K with $K \in \{1, 2, 5, 10, 100, 1000\}$ after augmentation and voting.

\subsection{Results}

\begin{table}[t]
\centering
\caption{ARC-AGI-1 Pass@K results. The hybrid (TR-mamba2attn) improves pass@2 (the official metric) by +2.0\% and consistently outperforms at higher K values, with the gap growing to +4.75\% at K=100, whilst maintaining pass@1 parity.}
\label{tab:arc_results}
\begin{tabular}{lcccccc}
\toprule
\textbf{K} & \textbf{1} & \textbf{2} & \textbf{5} & \textbf{10} & \textbf{100} & \textbf{1000} \\
\midrule
TRM-attn & 40.75 & 43.88 & 49.25 & 52.13 & 60.50 & 65.50 \\
TR-mamba2attn & 40.50 & 45.88 & 51.88 & 54.50 & 65.25 & 69.75 \\
\midrule
$\Delta$ & $-$0.25 & \textbf{+2.00} & \textbf{+2.63} & \textbf{+2.37} & \textbf{+4.75} & \textbf{+4.25} \\
\bottomrule
\end{tabular}
\end{table}

\begin{table}[t]
\centering
\caption{Sudoku-Extreme exact accuracy. MLP-t variants outperform attention-based models, with TR-mamba2mlpt slightly underperforming the TRM-mlp-t baseline, suggesting constraint satisfaction benefits from dense cross-position mixing.}
\label{tab:sudoku_results}
\begin{tabular}{lcc}
\toprule
\textbf{Model} & \textbf{Params} & \textbf{Accuracy (\%)} \\
\midrule
TRM-attn & 6.83M & 72.2 \\
TR-mamba2attn & 6.86M & 66.5 \\
\midrule
TRM-mlp-t & 5.00M & \textbf{87.4} \\
TR-mamba2mlpt & 6.28M & 84.2 \\
\bottomrule
\end{tabular}
\end{table}

\begin{table}[t]
\centering
\caption{Maze-30$\times$30-Hard exact accuracy. Training shows high variance across checkpoints for all models. Both MLP-t variants fail on this task, whilst TR-mamba2attn achieves 80.6\% at the final checkpoint, though results remain preliminary given instability.}
\label{tab:maze_results}
\begin{tabular}{lcc}
\toprule
\textbf{Model} & \textbf{Params} & \textbf{Accuracy (\%)} \\
\midrule
TRM-attn & 6.83M & 60.8 \\
TR-mamba2attn & 6.86M & 80.6 \\
\midrule
TRM-mlp-t & 19.0M & 0.0 \\
TR-mamba2mlpt & 13.24M & 0.0 \\
\bottomrule
\end{tabular}
\end{table}

\begin{figure}[t]
    \centering
    \includegraphics[width=\linewidth]{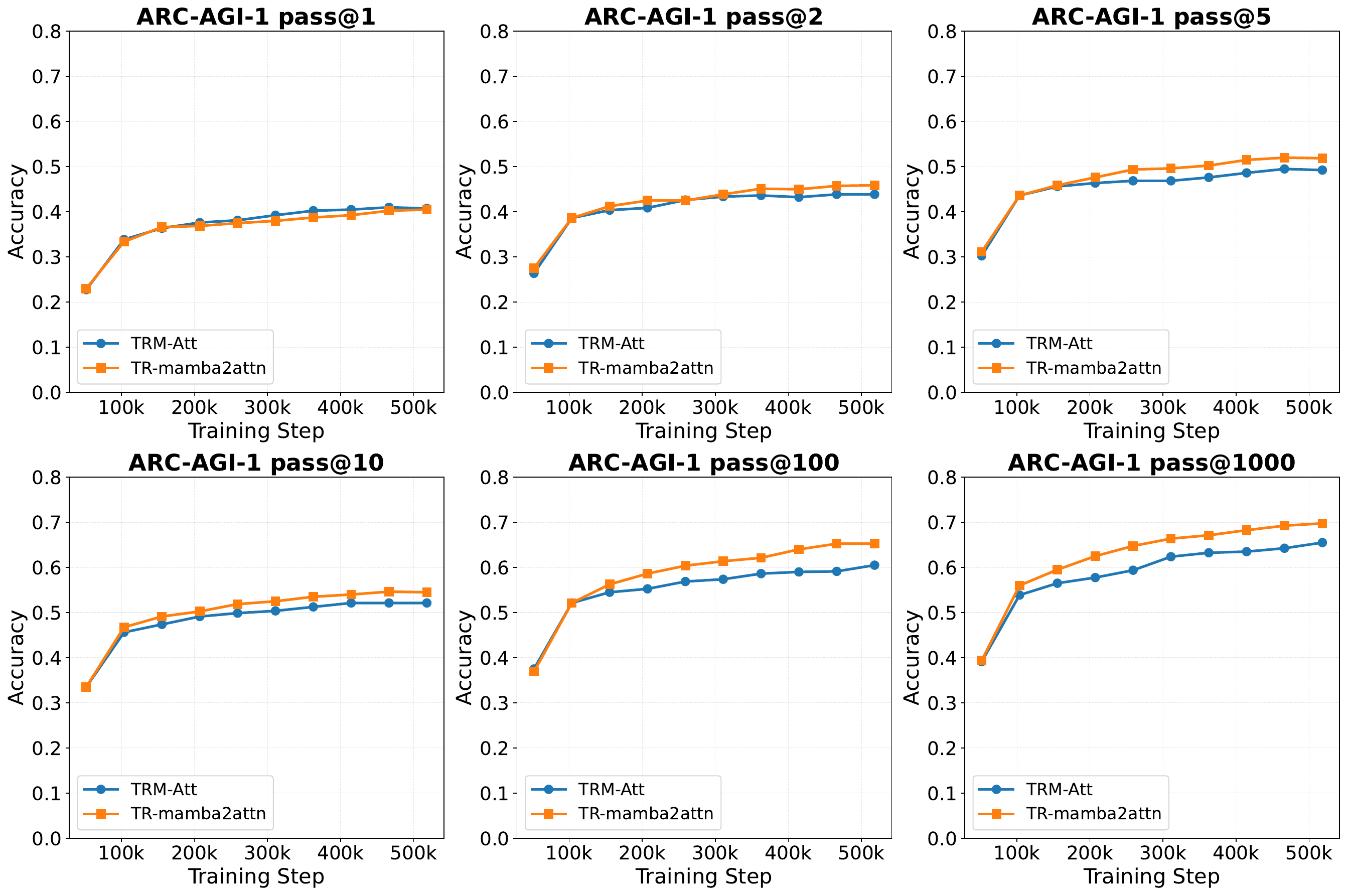}
    \caption{Training curves for ARC-AGI-1 across all pass@K metrics. The hybrid (TR-mamba2attn, orange) consistently outperforms the baseline (TRM-attn, blue) at pass@2 (the official metric) and higher K values throughout training, whilst maintaining pass@1 parity. The gap emerges early and remains stable, demonstrating that improved candidate coverage is a consistent property rather than a late-training phenomenon.}
    \label{fig:training_curves}
\end{figure}

\textbf{ARC-AGI-1} (Table~\ref{tab:arc_results}): TRM-mlp-t achieves 29.6\% pass@2 (19M parameters) as reported in the original TRM paper; we did not reproduce this result to conserve computational resources, instead focusing on the more capable attention-based TRM which warrants detailed investigation. TR-mamba2mlpt achieves 32.125\% pass@2 (13.24M parameters)---a +2.5\% improvement over the reported TRM-mlp-t baseline. We focus comparison on attention-based models. The hybrid (TR-mamba2attn) outperforms on pass@2 by +2.0\% (45.88\% vs 43.88\%), and the advantage grows at higher K values, reaching +4.75\% at K=100. The hybrid maintains near-parity at pass@1 ($-$0.25\%), indicating similar top-1 selection whilst improving candidate coverage. This pattern is consistent throughout training---the hybrid's pass@2 and pass@100 curves pull ahead of attention after the first few epochs and maintain the lead, as shown in Figure~\ref{fig:training_curves}.

\textbf{Sudoku} (Table~\ref{tab:sudoku_results}): The MLP-t variants achieve the strongest results, with TRM-mlp-t reaching 87.4\% (the best) and TR-mamba2mlpt achieving 84.2\%---both substantially outperforming attention-based models (TRM-attn: 72.2\%, TR-mamba2attn: 66.5\%). This suggests that constraint satisfaction on small, fixed grids (9$\times$9) benefits from dense all-to-all communication rather than selective attention or sequential processing.

\textbf{Maze} (Table~\ref{tab:maze_results}): In contrast to Sudoku, both MLP-t variants completely fail on the larger 30$\times$30 grids (0.0\% accuracy), whilst TR-mamba2attn achieves 80.6\% versus 60.8\% for TRM-attn. Training shows high variance across checkpoints (fluctuating between 6--85\% accuracy), making these results preliminary. The complete failure of MLP-t on this task highlights the importance of context-dependent architectural choices: dense mixing succeeds on small grids but fails to scale to larger spatial reasoning tasks.

\section{Discussion}

\subsection{Interpreting the Pass@K Pattern}

The ARC-AGI-1 evaluation protocol provides important context for interpreting pass@K metrics. The original test set contains 400 puzzles with 419 test inputs (19 puzzles have two test inputs, each scored independently), which are augmented through dihedral transformations and colour permutations to yield 368,150 total test instances---an average of $\sim$880$\times$ augmentation per test input. Each augmentation produces a prediction, which is then inverse-transformed and aggregated through voting. In this setting, pass@1000 effectively measures whether the correct answer appears \emph{anywhere} within the model's predicted candidate set across all augmentations. The hybrid model achieves 69.75\% at pass@1000 versus 65.50\% for TRM-attn (+4.25\%), indicating that the hybrid exhibits better \emph{coverage}: the correct solution is more likely to be generated somewhere within its diverse candidate pool.

The ARC results reveal a \emph{coverage vs selection} trade-off. The hybrid improves pass@2 by +2.0\%, and this advantage grows at higher K values:
\begin{itemize}
    \item \textbf{Coverage} (pass@2 and higher K): The hybrid generates the correct solution within its candidate set more often, improving candidate diversity.
    \item \textbf{Selection} (pass@1): Both models rank the correct solution first at similar rates (near-parity), so the hybrid's coverage gain does not come at the expense of top-1 selection.
\end{itemize}

\begin{figure}[t]
    \centering
    \includegraphics[width=0.9\linewidth]{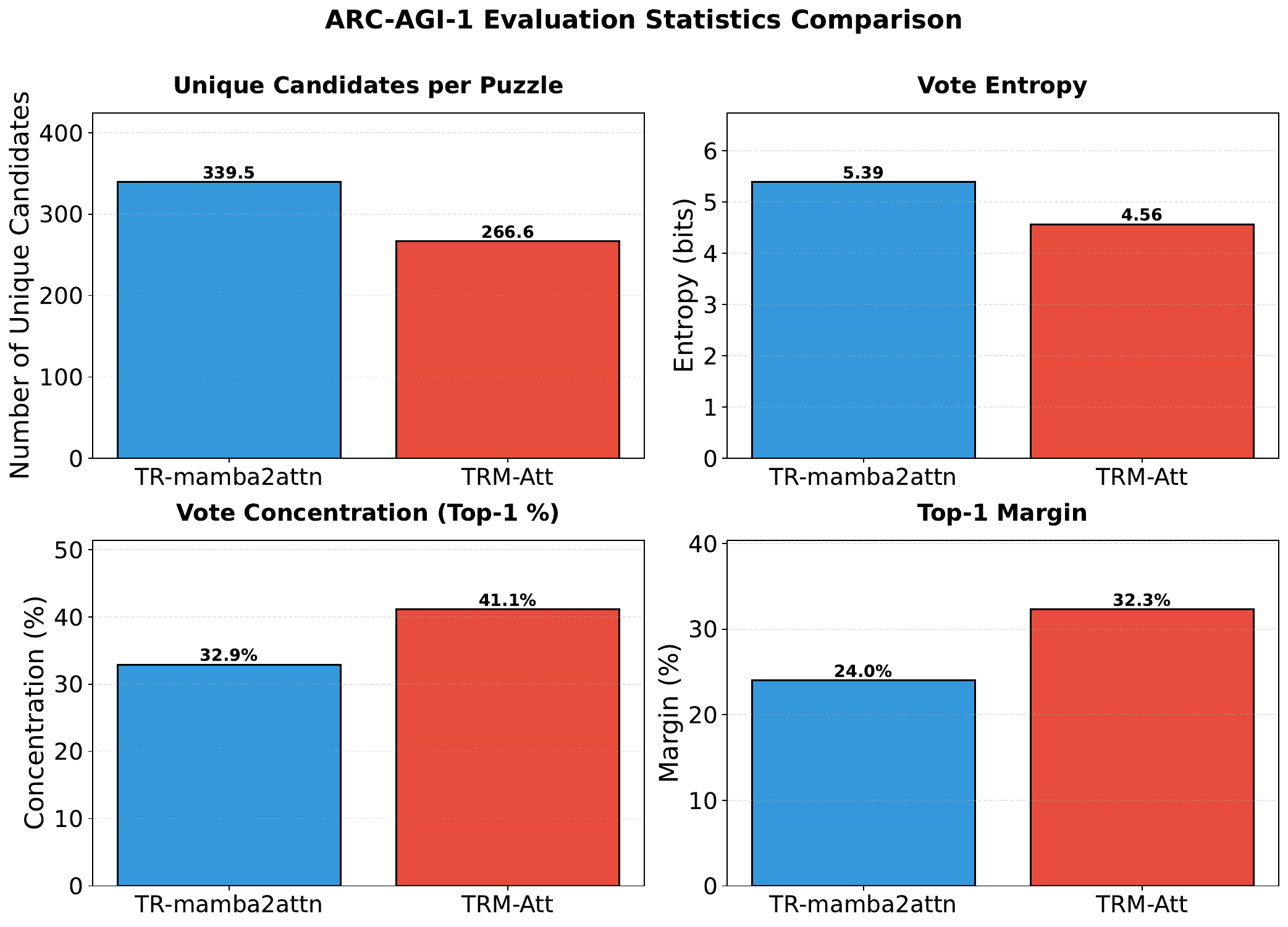}
    \caption{Evaluation statistics on ARC-AGI-1 comparing TR-mamba2attn (hybrid) and TRM-attn (baseline). The hybrid generates more unique candidates per puzzle and exhibits higher vote entropy (indicating diverse exploration), whilst TRM-attn shows higher vote concentration on the top-1 candidate and larger top-1 margin (indicating decisive selection). These statistics provide quantitative evidence for the coverage vs selection trade-off observed in pass@K curves.}
    \label{fig:eval_stats}
\end{figure}

To validate this hypothesis, we analyse the prediction statistics from the final checkpoints of both models on the ARC-AGI-1 evaluation set. Figure~\ref{fig:eval_stats} quantifies this trade-off through four key candidate metrics. The hybrid generates substantially more unique candidates per puzzle (339.5 vs 266.6, +27\%) with higher vote entropy (5.39 vs 4.56), indicating greater diversity in the candidate pool. Conversely, TRM-attn exhibits stronger selection: 41.1\% of votes concentrate on the top-1 candidate (vs 32.9\% for hybrid), with a larger top-1 margin (32.3\% vs 24.0\%). This explains the pass@K pattern: the hybrid's broader exploration yields more correct candidates within the pool (improving pass@K), whilst TRM-attn's more decisive voting concentrates on fewer high-confidence predictions (maintaining pass@1 parity). Mamba-2's sequential processing appears to contribute different solution trajectories during augmentation, increasing the diversity of the candidate pool without degrading the quality of the best prediction.

\subsection{Difficulty-Stratified Analysis}

Stratifying puzzles by difficulty sharpens this picture. We define difficulty using the average correct-vote share across both models (a model-agnostic measure): puzzles where neither model reliably produces the correct answer are ``hard,'' and those where correct answers dominate the candidate pool are ``easy.'' We stratify all 419 test inputs at a 15\% correct-vote-share threshold: hard ($<$15\%, $N$=246) and easy ($\geq$15\%, $N$=173). On hard test inputs, the hybrid gains +4.9\,pp at pass@5 over TRM-attn. The mechanism is intuitive: when the correct answer is rare in the candidate pool, the hybrid's flatter vote distribution avoids concentrating votes on a single dominant---but wrong---candidate, preserving the correct answer's chance of surfacing at higher K. On easy puzzles, the pattern reverses: TRM-attn gains +4.6\,pp at pass@1, because its sharper vote concentration more reliably promotes an already-dominant correct answer to the top rank. The aggregate pass@K pattern---near-parity at pass@1, growing hybrid advantage at higher K---is the net effect of these two opposing trends. At pass@5, the two models also solve partially disjoint puzzle sets (31 hybrid-only vs 23 TRM-attn-only), suggesting that different mixing strategies contribute complementary strengths.

\subsection{Mamba-2 Enters the Recursive Operator Design Space}

The key finding is that the hybrid excels at candidate coverage on ARC-AGI, improving pass@2 by +2.0\% with growing advantages at higher K values (+4--5\% at high K), whilst maintaining top-1 parity. Improved pass@2 performance on ARC alongside competitive results across Sudoku and Maze validates that introducing Mamba-2 into the recursive scaffold does not degrade reasoning capability---and on coverage metrics, improves it. This establishes Mamba-2 hybrid operators as viable candidates in the recursive operator design space, and motivates deeper investigation into the interaction between outer-loop recursion and SSMs' inherent inner recurrence.

    \section{Related Work}

    \textbf{Explicit CoT and token-space compute}: Chain-of-thought prompting \citep{wei2022cot} and self-consistency \citep{wang2022selfconsistency} scale reasoning by producing visible intermediate steps. A theoretical study proves that standard, bounded-depth Transformers are mathematically incapable of directly solving basic arithmetic or linear equations unless the model size grows super-polynomially \citep{feng2023cottheory}, providing theoretical grounding for why explicit reasoning steps are necessary. Tree of Thoughts \citep{yao2023tot} and scratchpads \citep{nye2021scratchpads} extend this to search and planning. We study recursive reasoning in latent space, which shares the same philosophy by extending the effective model depth through looping forward passes over the operator model.

    \textbf{Implicit/latent CoT}: Recent work connects chain-of-thought reasoning to latent reasoning by gradually converting explicit CoT models. \citet{deng2024stepwise} explore removing intermediate reasoning steps entirely or partially, though this approach generally lags behind explicit CoT in accuracy when no additional latent looping is provided. \citet{hao2024coconut} replace text tokens with continuous thought vectors, arguing that latent variables can encode multiple alternative future steps simultaneously. \citet{zelikman2024quietstar} use reinforcement learning to incentivise models to generate internal thoughts. More recently, \citet{yang2025trm, wang2025hierarchical} demonstrate that dual hidden states enable strong reasoning capability with tiny models (7M parameters) on specific reasoning benchmarks. Our work is directly built on TRM, replacing Transformer blocks with Mamba-2 hybrid operators and achieving improved ARC-AGI pass@K performance whilst maintaining competitive pass@1 accuracy.

    \textbf{Looped Transformers}: \citet{dehghani2018universal} first established the fundamental architecture for looped transformers with Universal Transformers. \citet{saunshi2025looped} explicitly study ``Looped Transformers'' denoted as $(k \otimes L)$, where a $k$-layer block is looped $L$ times, proving that these loops generate ``latent thoughts'' and can simulate chain-of-thought reasoning steps. \citet{yang2024looped} demonstrate that a looped transformer with 1 layer can match the performance of a standard 12-layer transformer by iterating the loop, focusing on emulating iterative algorithms like gradient descent. \citet{geiping2025recurrent} propose a depth-recurrent architecture featuring a core recurrent block sandwiched between a prelude and coda, where the core block is iterated $r$ times (randomly sampled during training) to allow the model to ``think'' in latent space. Although all these papers loop the operator model (transformer), input injection has been mentioned in multiple literature as a key design choice.

    \textbf{Efficient backbones}: State space models (SSMs) offer linear-time sequence modeling as an alternative to quadratic-complexity attention. \citet{gu2024mamba} introduce Mamba, which achieves linear-time performance through selective state spaces that dynamically modulate which information to retain or forget. \citet{dao2024mamba2} establish a theoretical duality between transformers and SSMs, showing that structured state space models can be viewed as a generalization of attention mechanisms, leading to Mamba-2 with improved efficiency and expressiveness. Hybrid architectures combining attention and SSMs have shown promise: \citet{lieber2024jamba} demonstrate that Jamba, a hybrid Transformer-Mamba model, achieves competitive performance at scale whilst maintaining efficiency benefits. More recently, \citet{zhang2025m1} explore Mamba's potential for test-time compute scaling, proposing M1 as a Mamba-based reasoning model that leverages the linear complexity to enable more intensive inference-time computation. Our work extends this direction by integrating Mamba-2 into recursive reasoning architectures, combining the efficiency of SSMs with the iterative refinement capabilities of looped transformers.

    \textbf{Normalisation in recursion}: Pre-norm vs post-norm affects training dynamics \citep{xiong2020prenorm}. DeepNet \citep{wang2022deepnet} and ReZero \citep{bachlechner2021rezero} address stability in deep networks. HRM \citep{wang2025hierarchical} specifically motivates post-norm for recursive Q-learning stability.

\section{Conclusion}

We investigated whether Mamba-2---whose state space recurrence is itself a form of iterative refinement---can enter the design space of operators for TRM-style recursive reasoning without degrading capability. By replacing Transformer blocks with Mamba-2 hybrid operators (parameter-matched), we found:
\begin{itemize}
    \item \textbf{ARC-AGI}: +2.0\% improvement on pass@2 (the official metric) with growing advantages at higher K values, suggesting better candidate coverage
    \item \textbf{Coverage-vs-selection trade-off}: Mamba-2's sequential processing contributes distinct solution trajectories, increasing candidate diversity without degrading top-1 quality. Difficulty stratification further shows the two operators solve partially disjoint puzzle sets, reinforcing that different mixing strategies bring complementary strengths within the recursive design space.
    \item \textbf{Post-norm}: Critical for stable recursive computation
\end{itemize}

Our results validate that Mamba-2 hybrid operators can enter the recursive operator design space with competitive performance, taking a firm first step towards understanding the best mixing strategies for recursive reasoning. Future work should investigate whether the recursive loop can be internalised into SSM state updates---leveraging Mamba's inherent inner recurrence---alongside compute-normalised evaluation.

\bibliography{references}

@article{wei2022cot,
  title={Chain-of-Thought Prompting Elicits Reasoning in Large Language Models},
  author={Wei, Jason and Wang, Xuezhi and Schuurmans, Dale and Bosma, Maarten and Ichter, Brian and Xia, Fei and Chi, Ed and Le, Quoc and Zhou, Denny},
  journal={Advances in Neural Information Processing Systems},
  volume={35},
  pages={24824--24837},
  year={2022}
}

@article{wang2022selfconsistency,
  title={Self-Consistency Improves Chain of Thought Reasoning in Language Models},
  author={Wang, Xuezhi and Wei, Jason and Schuurmans, Dale and Le, Quoc and Chi, Ed and Narang, Sharan and Chowdhery, Aakanksha and Zhou, Denny},
  journal={International Conference on Learning Representations},
  year={2022}
}

@article{yao2023tot,
  title={Tree of Thoughts: Deliberate Problem Solving with Large Language Models},
  author={Yao, Shunyu and Yu, Dian and Zhao, Jeffrey and Shafran, Izhak and Griffiths, Thomas L and Cao, Yuan and Narasimhan, Karthik},
  journal={Advances in Neural Information Processing Systems},
  volume={36},
  year={2023}
}

@article{nye2021scratchpads,
  title={Show Your Work: Scratchpads for Intermediate Computation with Language Models},
  author={Nye, Maxwell and Andreassen, Anders Johan and Gur-Ari, Guy and Michalewski, Henryk and Austin, Jacob and Biber, David and Dohan, David and Lewkowycz, Aitor and Bosma, Maarten and Luan, David and Sutton, Charles and Odena, Augustus},
  journal={arXiv preprint arXiv:2112.00114},
  year={2021}
}

@article{feng2023cottheory,
  title={Towards Revealing the Mystery behind Chain of Thought: A Theoretical Perspective},
  author={Feng, Guhao and Zhang, Bohang and Gu, Yuntian and Ye, Haotian and He, Di and Wang, Liwei},
  journal={Advances in Neural Information Processing Systems},
  volume={36},
  year={2023}
}

@article{lanham2023faithfulness,
  title={Measuring Faithfulness in Chain-of-Thought Reasoning},
  author={Lanham, Tamera and Chen, Anna and Radhakrishnan, Ansh and Steiner, Benoit and Denison, Carson and Hernandez, Danny and Li, Dustin and Durmus, Esin and Hubinger, Evan and Kernion, Jackson and others},
  journal={arXiv preprint arXiv:2307.13702},
  year={2023}
}

@article{deng2024stepwise,
  title={From Explicit CoT to Implicit CoT: Learning to Internalize CoT Step by Step},
  author={Deng, Yuntian and Choi, Yejin and Ritter, Samuel},
  journal={arXiv preprint arXiv:2405.14838},
  year={2024}
}

@article{hao2024coconut,
  title={Training Large Language Models to Reason in a Continuous Latent Space},
  author={Hao, Shibo and Gu, Sainbayar and Ma, Haotian and Hong, Joshua Jiahua and Wang, Zhen and Wang, Daisy Zhe and Hu, Zhiting},
  journal={arXiv preprint arXiv:2412.06769},
  year={2024}
}

@article{zelikman2024quietstar,
  title={Quiet-STaR: Language Models Can Teach Themselves to Think Before Speaking},
  author={Zelikman, Eric and Harik, Georges and Shao, Yijia and Jayasiri, Varuna and Haber, Nick and Goodman, Noah D},
  journal={Conference on Language Modeling},
  year={2024}
}

@article{dehghani2018universal,
  title={Universal Transformers},
  author={Dehghani, Mostafa and Gouws, Stephan and Vinyals, Oriol and Uszkoreit, Jakob and Kaiser, {\L}ukasz},
  journal={International Conference on Learning Representations},
  year={2019}
}

@article{geiping2025recurrent,
  title={Scaling up Test-Time Compute with Latent Reasoning: A Recurrent Depth Approach},
  author={Geiping, Jonas and McLeish, Sean and Jain, Neel and Kirchenbauer, John and Singh, Siddharth and Bartoldson, Brian R and Kailkhura, Bhavya and Bhatele, Abhinav and Goldstein, Tom},
  journal={arXiv preprint arXiv:2502.05171},
  year={2025}
}

@article{saunshi2025looped,
  title={Reasoning with Latent Thoughts: On the Power of Looped Transformers},
  author={Nikunj Saunshi and Nishanth Dikkala and Zhiyuan Li and Sanjiv Kumar and Sashank J. Reddi},
  journal={International Conference on Learning Representations},
  year={2025}
}

@article{yang2024looped,
  title={Looped Transformers are Better at Learning Learning Algorithms},
  author={Liu Yang and Kangwook Lee and Robert D Nowak and Dimitris Papailiopoulos},
  journal={International Conference on Learning Representations},
  year={2024}
}

@article{yang2025trm,
  title={Less is More: Recursive Reasoning with Tiny Networks},
  author={Jolicoeur-Martineau, Alexia},
  journal={arXiv preprint arXiv:2510.04871},
  year={2025}
}

@inproceedings{gu2024mamba,
  title={Mamba: Linear-Time Sequence Modeling with Selective State Spaces},
  author={Gu, Albert and Dao, Tri},
  booktitle={Conference on Language Modeling},
  year={2024}
}

@article{dao2024mamba2,
  title={Transformers are SSMs: Generalized Models and Efficient Algorithms Through Structured State Space Duality},
  author={Dao, Tri and Gu, Albert},
  journal={International conference on machine learning},
  year={2024}
}

@inproceedings{lieber2024jamba,
  title={Jamba: A Hybrid Transformer-Mamba Language Model},
  author={Barak Lenz and Opher Lieber and Alan Arazi and Amir Bergman and Avshalom Manevich and Barak Peleg and Ben Aviram and Chen Almagor and Clara Fridman and Dan Padnos and Daniel Gissin and Daniel Jannai and Dor Muhlgay and Dor Zimberg and Edden M. Gerber and Elad Dolev and Eran Krakovsky and Erez Safahi and Erez Schwartz and Gal Cohen and Gal Shachaf and Haim Rozenblum and Hofit Bata and Ido Blass and Inbal Magar and Itay Dalmedigos and Jhonathan Osin and Julie Fadlon and Maria Rozman and Matan Danos and Michael Gokhman and Mor Zusman and Naama Gidron and Nir Ratner and Noam Gat and Noam Rozen and Oded Fried and Ohad Leshno and Omer Antverg and Omri Abend and Or Dagan and Orit Cohavi and Raz Alon and Ro'i Belson and Roi Cohen and Rom Gilad and Roman Glozman and Shahar Lev and Shai Shalev-Shwartz and Shaked Haim Meirom and Tal Delbari and Tal Ness and Tomer Asida and Tom Ben Gal and Tom Braude and Uriya Pumerantz and Josh Cohen and Yonatan Belinkov and Yuval Globerson and Yuval Peleg Levy and Yoav Shoham},
  booktitle={International Conference on Learning Representations},
  year={2025}
}

@article{zhang2025m1,
  title={M1: Towards scalable test-time compute with mamba reasoning models},
  author={Wang, Junxiong and Li, Wen-Ding and Paliotta, Daniele and Ritter, Daniel and Rush, Alexander M and Dao, Tri},
  journal={arXiv preprint arXiv:2504.10449},
  year={2025}
}

@inproceedings{xiong2020prenorm,
  title={On Layer Normalization in the Transformer Architecture},
  author={Xiong, Ruibin and Yang, Yunchang and He, Di and Zheng, Kai and Zheng, Shuxin and Xing, Chen and Zhang, Huishuai and Lan, Yanyan and Wang, Liwei and Liu, Tie-Yan},
  booktitle={International Conference on Machine Learning},
  pages={10524--10533},
  year={2020},
  organization={PMLR}
}

@article{wang2022deepnet,
  title={DeepNet: Scaling Transformers to 1,000 Layers},
  author={Wang, Hongyu and Ma, Shuming and Dong, Li and Huang, Shaohan and Zhang, Dongdong and Wei, Furu},
  journal={arXiv preprint arXiv:2203.00555},
  year={2022}
}

@article{bachlechner2021rezero,
  title={ReZero is All You Need: Fast Convergence at Large Depth},
  author={Bachlechner, Thomas and Majumder, Bodhisattwa Prasad and Mao, Huanru Henry and Cottrell, Garrison W and McAuley, Julian},
  journal={arXiv preprint arXiv:2003.04887},
  year={2021}
}

@article{wang2025hierarchical,
  title={Hierarchical Reasoning Model},
  author={Wang, Guan and Li, Jin and Sun, Yuhao and Chen, Xing and Liu, Changling and Wu, Yue and Lu, Meng and Song, Sen and Yadkori, Yasin Abbasi},
  journal={arXiv preprint arXiv:2506.21734},
  year={2025}
}

@article{chollet2019arc,
  title={On the Measure of Intelligence},
  author={Chollet, Fran{\c{c}}ois},
  journal={arXiv preprint arXiv:1911.01547},
  year={2019}
}

@misc{arcprize2025,
  title={Announcing ARC-AGI-2 and ARC Prize 2025},
  author={ARC Prize Foundation},
  howpublished={\url{https://arcprize.org/blog/announcing-arc-agi-2-and-arc-prize-2025}},
  year={2025}
}
\bibliographystyle{iclr2026_conference}

\end{document}